\documentclass[conference]{IEEEtran}
\IEEEoverridecommandlockouts
\usepackage{cite}
\usepackage{amsmath,amssymb,amsfonts}
\usepackage{algorithmic}
\usepackage{textcomp}
\usepackage{xcolor}

\usepackage{amsmath,amsfonts,bm}

\def\eqref#1{equation~\ref{#1}}

\def\1{\bm{1}}

\DeclareMathAlphabet{\mathsfit}{\encodingdefault}{\sfdefault}{m}{sl}
\SetMathAlphabet{\mathsfit}{bold}{\encodingdefault}{\sfdefault}{bx}{n}

\newcommand{\E}{\mathbb{E}}

\usepackage{algorithm}
\usepackage{array}
\usepackage{CJK}
\usepackage{color}
\usepackage{diagbox}
\usepackage{epsfig}
\usepackage{graphicx}
\usepackage{indentfirst}
\usepackage{multirow} 
\usepackage[normalem]{ulem}
\usepackage{stmaryrd}
\usepackage{slashbox}
\usepackage{subfig}
\usepackage{times}
\usepackage{url}
\usepackage{mathtools}
\graphicspath{{./image/}}

\def\BibTeX{{\rm B\kern-.05em{\sc i\kern-.025em b}\kern-.08em
    T\kern-.1667em\lower.7ex\hbox{E}\kern-.125emX}}

\begin{document}

\title{On the Trend-corrected Variant of Adaptive Stochastic Optimization Methods\\
}

\author{\IEEEauthorblockN{ Bingxin Zhou*, Xuebin Zheng*}
\IEEEauthorblockA{\textit{The University of Sydney Business School} \\
\textit{The University of Sydney}\\
Sydney, Australia \\
\{bzho3923, xzhe2914\}@uni.sydney.edu.au \\
\textit{* equal contribution}}
\and
\IEEEauthorblockN{Junbin Gao}
\IEEEauthorblockA{\textit{The University of Sydney Business School} \\
\textit{The University of Sydney}\\
Sydney, Australia \\
junbin.gao@sydney.edu.au}
}

\maketitle

\begin{abstract}
Adam-type optimizers, as a class of adaptive moment estimation methods with the exponential moving average scheme, have been successfully used in many applications of deep learning. Such methods are appealing due to the capability on large-scale sparse datasets with high computational efficiency. In this paper, we present a new framework for Adam-type methods with the trend information when updating the parameters with the adaptive step size and gradients. The additional terms in the algorithm promise an efficient movement on the complex cost surface, and thus the loss would converge more rapidly. We show empirically the importance of adding the trend component, where our framework outperforms the conventional Adam and AMSGrad methods constantly on the classical models with several real-world datasets. 
\end{abstract}


\section{Introduction}
Employing first order optimization methods, such as Stochastic Gradient Descent (SGD), is the key of solving large-scale problems. The classic gradient descent algorithm is widely used to update the model parameters, denoted by $x$
\begin{equation}
x_{t+1}=x_{t}-\eta \nabla f(x_{t}),
\label{Proj134-1}
\end{equation}
where the gradient is denoted by $\nabla f(x_{t})$ and the step size by $\eta$. While the method has shown its efficiency for many contemporary tasks, the adaptive variants of SGD outperform the vanilla SGD methods on their rapid training time. Specifically, the step size $\eta$ is substituted by an adaptive step size $\eta / \sqrt{v_{t}}$, and $v_t$ is generated from the squared gradient $[\nabla f(x_{t})]^2$ where the operation is element-wise.

Several variants of the popular adaptive optimizers can be summarized into such common format. These optimizers share gradients calculation and parameters updating functions, but specify different moving average schemes for calculating the parameter-wise adaptive learning rate $\eta / \sqrt{v_{t}}$. For example, AdaGrad \cite{duchi2011adaptive} takes the arithmetic average of historical squared gradients $[\nabla f(x_{t})]^2$. Compared with the conventional momentum method, it adapts the learning rate to each parameter to suit the sparse data structure, and thus gains a rapid convergence speed \cite{ruder2016overview}. RMSProp \cite{tieleman2012lecture} was proposed to reduce the aggressiveness of the decay rate in AdaGrad . The method modifies $v_t$ to the exponentially decayed squared gradients. Similar implementations could also be found in ADADELTA \cite{zeiler2012adadelta}. Instead of the squared gradients, the method applies squared parameter updates to define the adaptive learning rate. As a result, each update guarantees the same hypothetical units as the parameter. Later, Adam \cite{kingma2014adam} modifies RMSProp with the idea from momentum methods \cite{qian1999momentum}. Except for the second moment moving average, the new rule also replaces the gradient $\nabla f(x_{t})$ at the end of (\ref{Proj134-1}) with the first-moment estimation. The method has practically shown its superiority regarding the converge speed and memory requirement. AMSGrad in \cite{reddi2018convergence} applies the maximum of past second-moment estimation to prevent converging to a suboptimal solutions where Adam could potentially trapped in. While the aforementioned methods are the most famous frameworks, many variants are also proposed in the last few years \cite{dozat2016incorporating, loshchilov2018decoupled, liu2020on}. In general, we call such adaptive methods with shared structures as Adam-type optimizers.

So far, the adaptive methods with exponential moving average gradients have gained great attention with huge success in many deep learning tasks. However, it remains unsolved whether the simple exponential smoothing results or the level information is sufficient in capturing the landscape of the cost surface. When clear upward or downward pattern could be recognized within the moving routine, it is suggested to add a trend term on top of the level-only information. 

In this paper, we propose the notion of trend corrected exponential smoothing to modify the conventional application of exponential moving average in optimizers. We name the \textbf{T}rend-corrected variant of Adam as \textit{AdamT} and the proposed method converges consistently as Adam at $\mathcal{O}(\sqrt{T})$. To the best of our knowledge, this research work is the first to apply the trend-corrected features on gradients scaling and parameters updating in the literature. We testify our framework on the vanilla Adam method for rule implementation and performance comparison. In addition, we provide supplementary evaluations on modified AMSGrad to avoid the potential suboptimal problems suffered by Adam. By using the same naming convention, we call the trend-corrected AMSGrad as \textit{AMSGradT}. The empirical results on some typical machine learning problems demonstrate the convergence and generalization ability of AdamT and AMSGradT in both convex and non-convex settings. It shall be emphasized that our framework is universally implementable for all adaptive update methods that involve the exponential moving average term, including but not restricted to ADADELTA, RMSProp, AdaMAX and other well-recognized methods. 

For the remainder of the paper, we present in Section~\ref{preliminary} the fundamental idea of Adam, AMSGrad and Holt's linear methods. In Sections~\ref{methodology} and \ref{experiment}, we detail the update rules of our proposed methods and experimental analysis respectively. In addition, Section~\ref{relatedWorK} reviews recent developments of Adam-type optimizers. While many of them focus more on non-convex optimizations, there is a potential to incorporate our methods with such frameworks and this extension is expected for future works.


\section{Preliminary}
\label{preliminary}
\subsection{Adaptive Methods with Exponential Moving Averages}
For adaptive gradient descent methods, the update rule can be written as
\[
x_{t+1} = x_{t}-\frac{\eta}{\sqrt{\hat{v}_{t}}+\epsilon} \hat{m}_t
\]
where $m_t$ is the gradient updates, and conventionally it is defined to be the last gradient value $\nabla f(x_{t})$. To prevent zero division, a smoothing term $\epsilon$ is included in the denominator. 

We initially focus our analysis and modifications on Adam \cite{kingma2014adam}, one of the most popular optimizers in the last few years. The adaptive step size involves the squared gradients
\begin{equation}
v_t = \alpha v_{t-1}+(1-\alpha) \nabla f(x_t) \circ \nabla f(x_t)
\label{adam_v}
\end{equation}
where the operation ``$\circ$'' denotes an element-wise multiplication. In terms of the gradient $m_t$, Adam takes the exponentially weighted average of all previous gradients instead of solely relying on the last gradient value $\nabla f(x_{t})$
\begin{equation}
m_{t} = \beta_1 m_{t-1}+\left(1-\beta_1\right) \nabla f(x_{t}).
\label{adam_m}
\end{equation}

While the two moment estimates from (\ref{adam_v}) \& (\ref{adam_m}) could potentially counteract towards zero, the series $\hat{m}_{t}$ and $\hat{v}_{t}$ are considered for bias-correction. Formally, the rules are defined as:
$$
\begin{aligned} 
\hat{m}_{t} =\frac{m_{t}}{1-\beta_{1}^{t}}; \
\quad\hat{v}_{t} =\frac{v_{t}}{1-\beta_{2}^{t}}.
\end{aligned}
$$

To allow rare mini-batches provide informative gradients for a promising optimal solution, AMSGrad \cite{reddi2018convergence} includes long-term memory in squared gradients. Instead of using the latest $v_t$, AMSGrad employs the following update rule for the second moment estimate
$$
\begin{aligned}
\quad v_{t}^{\textnormal{max}} = \max(v_{t-1}^{\textnormal{max}}, {v}_{t}),
\end{aligned}
$$
and then use $v_{t}^{\textnormal{max}}$ to make the learning rate adaptive. As a result, a non-increasing step size is utilized during the updating process, which avoids the problems suffered from the conventional adaptive learning methods, including Adam. 

\subsection{Trend Corrected Exponential Smoothing}
\label{holts}
The idea of extracting a smoothed new point from all the previous information is called the exponential weighted moving average. The method was extended in \cite{holt2004forecasting} by including the trend behaviours within the series, namely trend-corrected exponential smoothing or Holt's linear method. Consider a time series \{$y_t$\} for $t = 1, 2, \ldots, T$, our target is to find the smoothing results $\widehat{y_{t+1|t}}$. Holt's linear method formulates the conditional forecasting by summing up two exponential smoothing equations
\begin{eqnarray}
\widehat{y_{t+1 | t}} 
&=& \ell_{t}+b_{t} \nonumber\\
\ell_{t} 
&=& \beta(\ell_{t-1}+b_{t-1})+(1-\beta)y_{t} \nonumber\\
b_{t} 
&=& \gamma b_{t-1}+(1-\gamma)(\ell_{t}-\ell_{t-1}). \nonumber
\end{eqnarray}

For a new estimation, we first update the level term $\ell_t$ with the weighted average of the last observation $y_t$ and its estimation $\widehat{y_{t|t-1}}$. The trend term $b_t$ is updated afterwards as the weighted average of the estimated trend $\ell_t - \ell_{t-1}$ and its previous estimation $b_{t-1}$. The smoothing parameters for the level and the trend are denoted as $\beta$ and $\gamma$ respectively. Both values could be selected between $0$ and $1$. 

Including a damping factor $\phi$ is also suggested in \cite{gardner1985forecasting}, so that the Holt's linear method with damping factor has the following form
\begin{eqnarray}
\widehat{y_{t+1|t}} &=& \ell_{t}+\phi b_{t} \nonumber\\
\ell_{t} &=& \beta(\ell_{t-1}+\phi b_{t-1})+(1-\beta) y_{t} \nonumber\\ 
b_{t} &=& \gamma\phi b_{t-1}+(1-\gamma)(\ell_{t}-\ell_{t-1}). \nonumber
\end{eqnarray}

The damped method will be identical to Holt's linear method if $\phi = 1$, and will be the same as simple exponential moving average method if we set $\phi = 0$. When $\phi$ is restricted to be positive, the damping factor could be used to control the significance of the trend component. 

The damped trend methods are considerably popular for forecasting tasks \cite{hyndman2018forecasting}. Such methods inherent both level and trend information from historical series, while staying flexible enough to adjust the influence of the trend term via $\phi$. On top of that, involving the damped factor could to some extend reduce the volatility of the smoothed line.


\section{Methodology}\label{methodology}
We introduce the trend-corrected variants of two adam-type stochastic optimization methods, namely \textit{AdamT} and \textit{AMSGradT}. The proposed methods are based on \cite{kingma2014adam} and \cite{reddi2018convergence} with added Holt's linear trend information for both of the first moment estimate and the second raw moment estimate. Specifically, we use trend-corrected 
exponential weighted moving averages in the final parameter update step instead of the level-only estimates used in Adam and AMSGrad.

\subsection{Algorithm for AdamT}
Consider the gradient of a stochastic objective function $f(x)$ evaluated at $T$ iterations as a time series $\{\nabla f(x_t)\}$ for $t=1,2,\ldots,T$. According to the Holt's linear trend method illustrated in Section~\ref{holts}, we write two series $\{\ell_t^m\}$ and $\{b_t^m\}$ as the exponential weighted moving averages which estimate the level and trend information of the first moment $\nabla f(x_t)$:
\begin{eqnarray}
    \ell_{t}^{m} &=&\beta_{1}m_{t-1}+(1-\beta_{1}) \nabla f(x_t) \label{extra_0}\\
	b_{t}^{m} &=&\gamma_{1}\phi_1 b_{t-1}^{m}+(1-\gamma_{1}) (\ell_{t}^{m}-\ell_{t-1}^{m})\label{extra_1} \\
	m_t&=&\ell_t^m+\phi_1b_t^m \label{m_t}
\end{eqnarray}
where $\beta_1, \gamma_1$ and $\phi_1$ have the same functionality as explained in Section~\ref{holts} and these are regarded as hyperparameters in our algorithm. Equation~(\ref{m_t}) combines the level and the trend information of first moment, which will be used for calculating the final update rule and the trend-corrected level estimates. The procedures for the second raw moment $\nabla f(x_t) \circ \nabla f(x_t)$ is analogous:
\begin{eqnarray}
	\ell_{t}^{v} &=& \beta_{2}v_{t-1}+(1-\beta_{2}) \nabla f(x_t) \circ \nabla f(x_t) \label{l_v}\\
	b_{t}^{v} &=& \gamma_{2}\phi_2 b_{t-1}^{v}+(1-\gamma_{2})(\ell_{t}^{v}-\ell_{t-1}^{v}) \label{b_v} \\
	v_t &=& \ell_t^v+\phi_2b_t^v \label{v_t}
\end{eqnarray}
The hyperparameters $\beta_2, \gamma_2$ and $\phi_2$ here share the same corresponding meanings as before. The moving averages $\{\ell_t^v\}$ and $\{b_t^v\}$ estimate the level and trend of the second raw moment respectively. The term $v_t$ combines these two information, which will be used in the calculations of final update rule and trend-corrected level estimates of the second raw moment.

In our algorithm, we set the initial values of the series $\{\ell^m_t\}, \{\ell^v_t\}, \{b^m_t\}$ and $\{b^v_t\}$ to be zero vectors, so that $\ell^m_0=\ell^v_0=b^m_0=b^v_0=0$. The series $\{m_t\}$ and $\{v_t\}$, as a result, are also initialized as zero vectors. As observed in \cite{kingma2014adam}, the exponential weighted moving averages could bias towards zero, especially during the early training stage. We perform the bias correction for the two level estimates $\{\ell_t^m\}$ and $\{\ell_t^v\}$ by following \cite{kingma2014adam}. For the two trend estimates $\{b_t^m\}$ and $\{b_t^v\}$, we correct the bias in a different way by taking into account the effect of damping parameters ($\phi_1, \phi_2$). Thus, the bias-corrected version of the series $\{m_t\}$ and $\{v_t\}$ can be written as:
\begin{align}
    \hat{m}_t&=\frac{\ell_t^m}{1-\beta_1^t}+\frac{(1-\gamma_1\phi_1)b_t^m}{(1-\gamma_1)(1-(\gamma_1\phi_1)^t)} \label{m_hat}\\
    \quad\hat{v}_t&=\frac{\ell_t^v}{1-\beta_2^t}+\frac{(1-\gamma_2\phi_2)b_t^v}{(1-\gamma_2)(1-(\gamma_2\phi_2)^t)} \label{v_hat}.
\end{align}

The justification for the two bias-corrected trend estimates $\{b_t^m\}$ and $\{b_t^v\}$ is provided below, where we have to take into account the effect of the corresponding damping factors. Here, we give the justification for the trend estimate $\{b_t^m\}$, and the procedures for $\{b_t^v\}$ is analogous. Note that we can write the trend estimate $b_t^m$ into the following compact summation form:
\begin{align*}
    b_{t}^{m} &=\gamma_{1}\phi_1 b_{t-1}^{m}+(1-\gamma_{1}) (\ell_{t}^{m}-\ell_{t-1}^{m})\\
    &= (1-\gamma_1)\sum_{i=1}^t (\gamma_1\phi_1)^{t-i}(\ell_i^m-\ell_{i-1}^m).
\end{align*}

To find how the expectation of the trend estimates $b_t^m$ relates to the expectation of the difference between the level estimates at successive timesteps $(\ell_t^m-\ell_{t-1}^m)$, we take the expectation for both sides of the above equation:
\begin{align*}
    \E[b_{t}^{m}] &= \E[(1-\gamma_1)\sum_{i=1}^t (\gamma_1\phi_1)^{t-i}(\ell_i^m-\ell_{i-1}^m)]\\
    &=(1-\gamma_1)\sum_{i=1}^t (\gamma_1\phi_1)^{t-i}\E[(\ell_i^m-\ell_{i-1}^m)]\\
    &=\E[(\ell_t^m-\ell_{t-1}^m)](1-\gamma_1)\sum_{i=1}^t (\gamma_1\phi_1)^{t-i}+\zeta,
\end{align*}
where $\zeta$ can be considered as a small constant, since the factor $(\gamma_1\phi_1)^{t-i}$ will be tiny if the associated expectation $\E[(\ell_i^m-\ell_{i-1}^m)]$ is too far away in the past in the case that $\E[(\ell_i^m-\ell_{i-1}^m)]$ is non-stationary. If $\E[(\ell_i^m-\ell_{i-1}^m)]$ is stationary, the constant $\zeta$ will be zero. To further simplify the above equation, we apply the formula for the sum of geometric sequence:
\[
\E[b_{t}^{m}] = \E[(\ell_t^m-\ell_{t-1}^m)](1-\gamma_1)\left(\frac{1-(\gamma_1\phi_1)^t}{1-\gamma_1\phi_1}\right)+\zeta.
\]

This suggests that we can use the term $(1-\gamma_1)[1-(\gamma_1\phi_1)^t]/[1-\gamma_1\phi_1]$ to correct the bias and close the discrepancy between the above two expectations at the presence of the damping factor $\phi_1$.

The final adaptive update rule is similar to Adam with the bias-corrected first moment estimate and the second raw moment estimate:
\begin{equation}
    x_{t+1} = x_{t}-\frac{\eta}{\sqrt{\lvert\hat{v}_t\rvert}+\epsilon}\hat{m}_t \label{update_rule}
\end{equation}
where $\epsilon$ is a positive tiny number added in the denominator to avoid zero-division case. Please note that the series $\{\hat{m}_t\}$ and $\{\hat{v}_t\}$ in AdamT are different from that of Adam. The two series are trend-corrected (also bias-corrected) estimates of both moments. Also, we use the absolute value of the series $\{\hat{v}_t\}$ under the square root in the denominator due to the possible negative values from the series $\{\hat{v}_t\}$.

The direction of the effective step $\Delta_t=\eta\cdot\hat{m}_t/\sqrt{\lvert\hat{v}_t\rvert}$ (with $\epsilon=0$) in the parameter space depends on the joint effect of the first moment level and trend estimates. In the update rule~(\ref{update_rule}), we only care about the magnitude of $\hat{v}_t$ by taking the absolute value and thus the ratio $\hat{m}_t/\sqrt{\lvert\hat{v}_t\rvert}$ can be seen as a signal-to-noise ratio. Note that the effective step $\Delta_t$ in our algorithm is also invariant to the scale of the gradients. Specifically, re-scaling the gradients $\nabla f(x_t)$ with a factor $c$ will scale $\ell_t^m$ and $b_t^m$ by a factor $c$, and will scale $\ell_t^v$ and $b_t^v$ by a factor $c^2$. This results in scaling $\hat{m}_t$ and $\hat{v}_t$ by a factor $c$ and $c^2$ respectively, and finally cancel out in the parameter update rule $(c\cdot\hat{m}_t)/(\sqrt{\lvert c^2\cdot\hat{v}_t\rvert})=\hat{m}_t/\sqrt{\lvert\hat{v}_t\rvert}$.

Note that our proposed method AdamT has two extra computational steps, that is (\ref{extra_1}) \& (\ref{m_t}). However, the computational complexity of these two steps is almost linear in time. Therefore, we can conclude that AdamT yields a superior performance compared with Adam (the results will be shown in the experiment section) with a minimal additional computational cost.

In our algorithm, we set the hyperparameters $\beta_1, \gamma_1, \beta_2, \gamma_2$ according to the suggestions in \cite{kingma2014adam}. The smoothing parameters for the first moment estimates are set to $0.9$, that is $\beta_1=\gamma_1=0.9$, while the smoothing parameters for the second raw moment estimates are set to $0.999$, that is $\beta_2=\gamma_2=0.999$. We empirically find that the good default values of the two damping parameters can be set to $\phi_1=\phi_2=0.5$. 
The pseudo-code of our AdamT is provided in Algorithm~\ref{AdamT}. 


\begin{algorithm}
\caption{The \textbf{Adam} optimizer modified with Holt's Linear \textbf{T}rend method. Empirically suggested default values for the hyperparameters are $\beta_1=\gamma_1=0.9, \beta_2=\gamma_2=0.999, \phi_1=\phi_2=0.5, \eta=0.0001$. All of the operations on vectors are element-wise.}
\label{AdamT}
\begin{algorithmic}[1]
\renewcommand{\algorithmicrequire}{\textbf{Input:}}
\renewcommand{\algorithmicensure}{\textbf{Output:}}
\REQUIRE Learning rate $\eta$; Smoothing factors $\beta_1$, $\gamma_1$, $\beta_2$, $\gamma_2$; Damping factors $\phi_1, \phi_2$; Noisy objective function $f(x)$ with parameters $x$
\ENSURE  Optimal model parameters $x^*$ 
\STATE \textit{Initialization}\\
    \hspace{0.2cm} $x_1$: Initial parameter values\\
    \hspace{0.2cm} $\ell_0^m\leftarrow0$: Initial first moment level estimate\\
    \hspace{0.2cm} $\ell_0^v\leftarrow0$: Initial second raw moment level estimate\\
    \hspace{0.2cm} $b_0^m\leftarrow0$: Initial first moment trend estimate\\
    \hspace{0.2cm} $b_0^v\leftarrow0$: Initial second raw moment trend estimate
\FOR {$t = 1$ to $T$}
    \STATE $\ell_{t}^{m} \leftarrow \beta_{1}m_{t-1}+(1-\beta_{1}) \nabla f(x_t)$
    \STATE $b_{t}^{m} \leftarrow \gamma_{1}\phi_1 b_{t-1}^{m}+(1-\gamma_{1}) (\ell_{t}^{m}-\ell_{t-1}^{m})$
    \STATE $\ell_{t}^{v} \leftarrow \beta_{2}v_{t-1}+(1-\beta_{2}) \nabla f(x_t) \circ \nabla f(x_t)$
    \STATE $b_{t}^{v} \leftarrow \gamma_{2}\phi_2 b_{t-1}^{v}+(1-\gamma_{2})(\ell_{t}^{v}-\ell_{t-1}^{v})$
    \STATE $\hat{m}_t \leftarrow \ell_t^m/(1-\beta_1^t)+[(1-\gamma_1\phi_1)b_t^m]/ [(1-\gamma_1)(1-(\gamma_1\phi_1)^t)]$
    \STATE $\hat{v}_t \leftarrow \ell_t^v/(1-\beta_2^t)+[(1-\gamma_2\phi_2)b_t^v]/ [(1-\gamma_2)(1-(\gamma_2\phi_2)^t)]$
    \STATE $x_{t+1} \leftarrow x_{t}-\eta\hat{m}_t/(\sqrt{\lvert\hat{v}_t\rvert}+\epsilon)$
\ENDFOR
\RETURN $x^*$ 
\end{algorithmic} 
\end{algorithm}

\subsection{Algorithm for AMSGradT}
We introduce here the variant AMSGradT to overcome the non-convergence and suboptimal problems of the proposed AdamT by using the same technique illustrated in \cite{reddi2018convergence}. These issues of AdamT are inherited from the base approach Adam \cite{kingma2014adam}. Essentially, we follow the same procedures in (\ref{extra_0})-(\ref{m_t}) and (\ref{m_hat}) for the first moment estimation with bias-correction. For the second raw moment estimate (\ref{l_v})-(\ref{v_t}), we add one more step after (\ref{v_t}):
\begin{align*}
v_{t}^{\max} = \max(v_{t-1}^{\max}, v_{t}),
\end{align*}
where $v_t$ is the second raw moment estimation from (\ref{v_t}) before bias correction. We set the initial value of the series $\{v_{t}^{\max}\}$ to be $v_{0}^{\max}=0$. As a result, the new bias-correction step which is used to replace (\ref{v_hat}) can be written as
\begin{align*}
\hat{v}_{t}^{\max} = \frac{(\ell_t^v)^{\max}}{1-\beta_2^t} + \frac{(1-\gamma_2\phi_2)({b_t^v})^{\max}}{ (1-\gamma_2)(1-(\gamma_2\phi_2)^t)},
\end{align*}
where $(\ell_t^v)^{\max}$ and $(b_t^v)^{\max}$ are the corresponding level and trend information that used to calculate $v_t^{\max}$. Finally, we replace $\hat{v}_t$ in the final update rule (\ref{update_rule}) with the bias-corrected estimates $\hat{v}_t^{\max}$.


\section{Experiments}\label{experiment}
We evaluate the two proposed algorithms \textit{AdamT} and \textit{AMSGradT} on both convex and non-convex real-world optimization problems with several popular types of machine learning models. The models we considered in the experiments include logistic regression which has a well-known convex loss surface, and different neural network models, including feedforward neural networks, convolutional neural networks and variational autoencoder. Neural Networks with non-linear activation function typically have an inherent non-convex loss surface which is more challenging for an optimization method.

We compare our methods with the baseline approaches Adam \cite{kingma2014adam} and AMSGrad \cite{reddi2018convergence}, and then demonstrate the effectiveness of the trend information of the gradients infused in our proposed algorithms. The experiment results show that our methods AdamT and AMSGradT converge more quickly and reach a better minimum point than Adam and AMSGrad respectively. The observation evidences that the added trend information effectively helps AdamT and AMSGradT to better capture the landscape of loss surface. 

In each of the following experiments, we use the same set of initial values for the models, so that the initial model losses (the loss value at $\textnormal{epoch} = 0$) are identical for all the optimization methods. In terms of the hyperparameters, all the smoothing parameters ($\beta_1, \beta_2$ in Adam \& AMSGrad and $\beta_1, \beta_2, \gamma_1, \gamma_2$ in AdamT \& AMSGradT) are set at their corresponding default values which are provided in Algorithm~\ref{AdamT}. The damping factors $(\phi_1, \phi_2)$ are searched within the range $[0.1, 1.0)$ and the learning rate $\eta$ is also tuned through a grid search $\{1e-4, 5e-4, 1e-3, 5e-3\}$ to produce the best results for all of the optimizers. All the experiments and optimizers are written in PyTorch and the implementations of our proposed optimizers can be found at \textcolor{magenta}{https://github.com/xuebin-zh/AdamT}.

For the sake of saving space, we only visualize the convergence of Adam and AdamT in the paper for comparisons. The relative convergence performance of AMSGrad and AMSGradT is almost the same in the corresponding experiments.

\subsection{Logistic Regression for Fashion-MNIST}
We first evaluate our methods on the logistic regression (LR) for multi-class classification problem with Fashion-MNIST dataset \cite{xiao2017fashion} which is a MNIST-like dataset of fashion products. The dataset has $60,000$ training samples and $10,000$ testing samples. Each of them has $28\times28$ pixels. Each of the samples is classified into one of the $10$ fashion products. The cross-entropy loss function has a well-behaved convex surface. The learning rate $\eta$ is set to be constant during the training procedure. We use minibatch training with size set to $128$. 

\begin{figure}[t]
    \centerline{\includegraphics[height=2.4in]{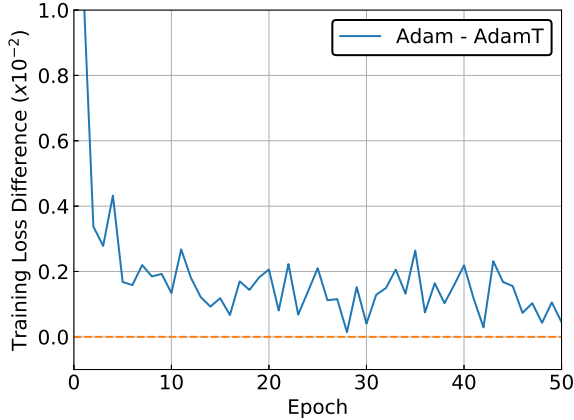}}
        \caption{Training loss difference of the logistic regression on Fashion-MNIST dataset for classification task.} 
    \label{exp_LR} 
\end{figure}

The training results are reported in Fig.~\ref{exp_LR}. Since the superiority of our method over Adam is relatively small in this experiment, the plot of loss value against epoch cannot visualize the difference. Instead, we plot the loss difference of the two optimizers, which is ($Loss_{\textnormal{Adam}}-Loss_{\textnormal{AdamT}}$) against training epoch. The difference above zero reflect the advantage of our AdamT. Fig.~\ref{exp_LR} indicates that AdamT converges faster at the early training stage and constantly outperforms Adam during the rest of the training phase, though the advantage is relatively small in this experiment. The loss surface of logistic regression is convex and well-behaved so that the trend information of AdamT cannot further provide much useful information for optimization, which results in a small advantage in this experiment.

\subsection{Feedforward Neural Networks for SVHN}
To investigate the performance on non-convex objective functions, we conduct the experiment with feedforward neural networks on \textit{The Street View House Numbers} (SVHN) dataset \cite{netzer2011reading} for a digit classification problem. We pre-process this RGB image dataset into grayscale for dimension reduction by taking the average across the channels for each pixel in the image. The samples are $32\times32$ grayscale images. The neural network used in this experiment has two fully-connected hidden layers, each of which has $1,400$ hidden units and ReLU activation function is used for the two hidden layers. The softmax cross-entropy loss function is used for training.

To evaluate the performance of the optimizers in noisy settings, we apply a stochastic regularization method in the model for a separate experiment. Specifically, we include two dropout layers \cite{srivastava2014dropout}, where one is applied between the two hidden layers and the other one is used before the output layer. The dropout probability is set to $0.5$ for both of the two dropout layers. In the experiments, we use a constant learning rate $\eta$ and minibatch training with size set to $128$.

\begin{figure*}
    \centering
        \includegraphics[height=2.4in]{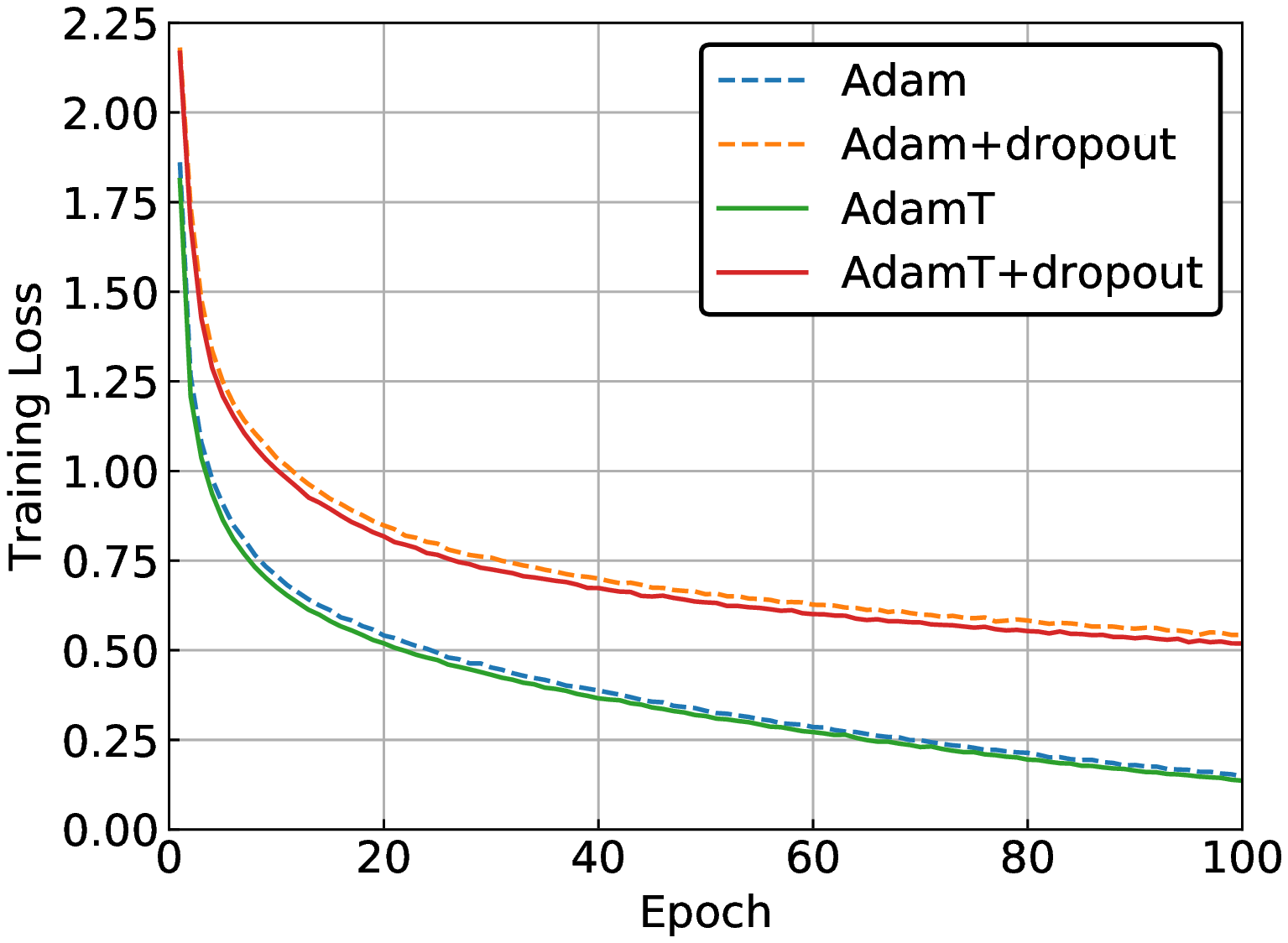}
        \includegraphics[height=2.4in]{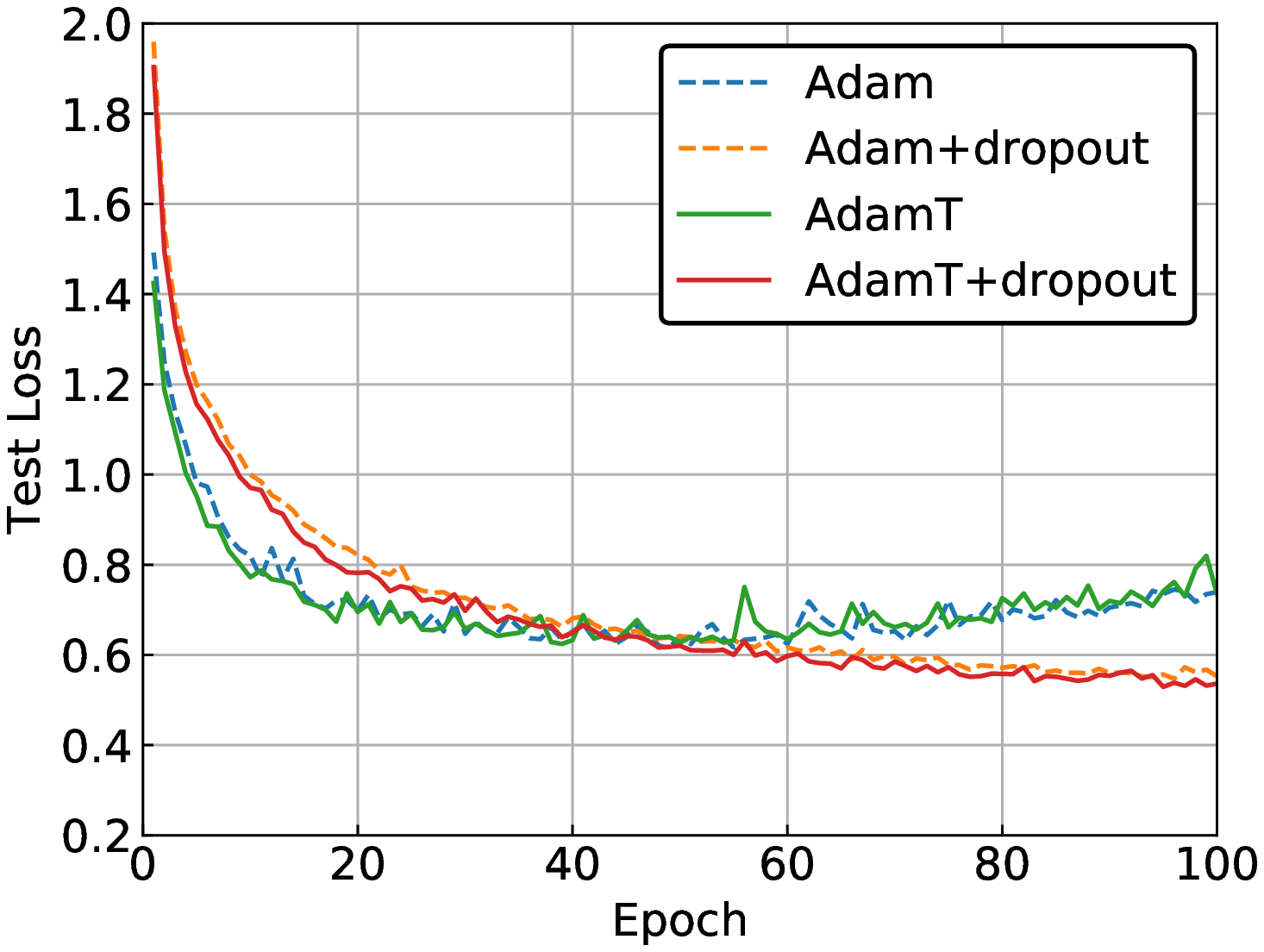}
    \caption{Training loss (left) and test loss (right) of the feedforward neural network on SVHN dataset for a classification problem. The model architecture is fc$1400$-fc$1400$.} 
    \label{exp_mnn}
\end{figure*}

We examine the convergence with and without dropout layers for the two optimizers. According to Fig.~\ref{exp_mnn}, we find that AdamT outperforms Adam obviously. In terms of the training process, AdamT yields a faster convergence and reaches a better position than Adam for the models, both with and without dropout layers. The superior performance of AdamT is also shown in the test phase, which demonstrates that AdamT also has a better generalization ability than Adam. For the model without dropout, the test results show that the model is prone to over-fitting and our method performs on a par with Adam. Comparing to logistic regression, the loss surface in this experiment becomes complex and non-convex. The trend estimates of the gradients from AdamT can provide more meaningful information of the landscape of the loss surface, and it encourages a better performance on the AdamT.

\subsection{Convolutional Neural Networks for CIFAR-10}
Convolutional neural network (CNN) is the main workhorse for Computer Vision tasks. We train a CNN model on standard CIFAR-10 dataset for a multi-class classification task. The dataset contains $50,000$ training samples and $10,000$ test samples, and each sample is an RGB $32\times32$ image. We pre-process the dataset by normalizing the pixel values to the range $[-1, 1]$ for a more robust training. The CNN model employed in this experiment is similar to the model used in \cite{reddi2018convergence}, which has the following architecture. There are $2$ stages of alternating convolution and max pooling layers. Each convolution layer has $64$ channels and kernel size $5\times 5$ with stride $1$. Each max pooling layer is applied with a kernel size of $2\times 2$. After that, there is a fully-connected layer with $600$ hidden units and a dropout probability $0.5$ followed by the output layer with $10$ units. We use ReLU for the activation function and softmax cross-entropy for the loss function. The model is trained with a tuned constant learning rate and minibatch size $128$ same as the previous experiments.

\begin{figure*}
    \centering
        \includegraphics[height=2.4in]{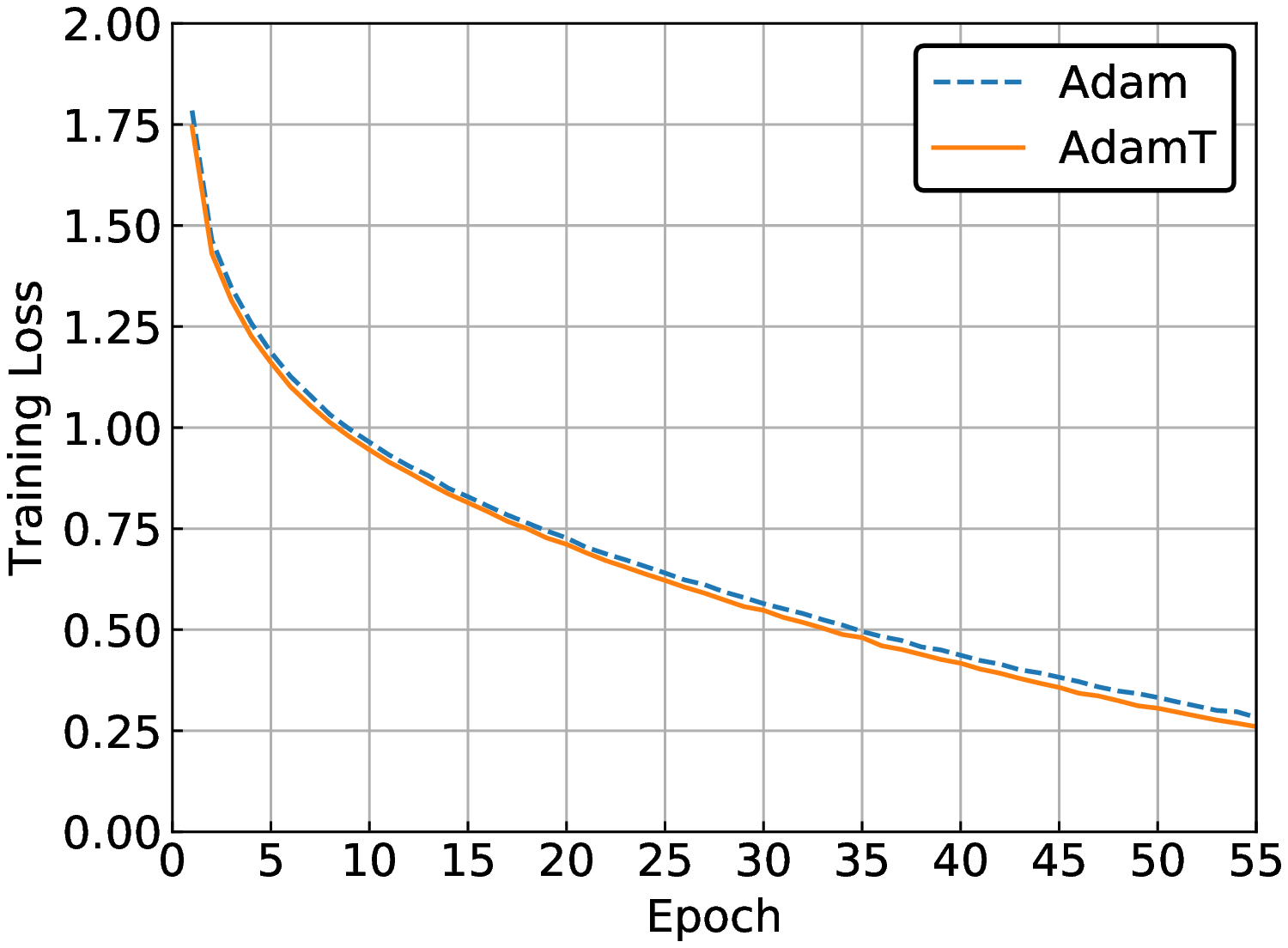}
        \includegraphics[height=2.4in]{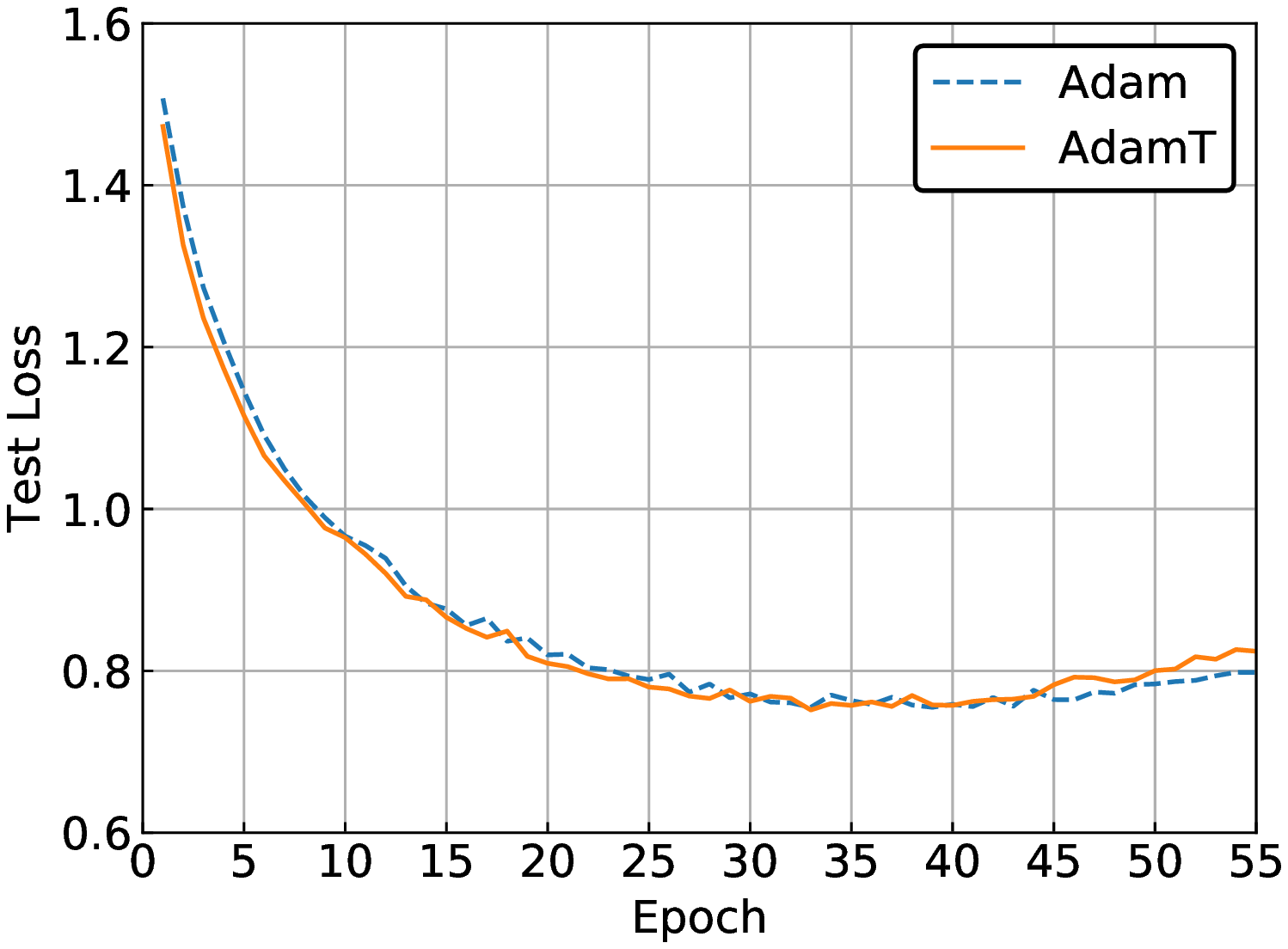}
    \caption{Training loss (left) and test loss (right) of the convolutional neural network on CIFAR-10 dataset for a classification task. The model architecture is c$64$-c$64$-fc$600$.} 
    \label{exp_cnn} 
\end{figure*}

We evaluate the test loss after each epoch during the training procedure and cease the training once the test loss of the model starts to increase. The training curves are reported in Fig.~\ref{exp_cnn}. We can observe that the proposed AdamT clearly excels Adam on the training loss, and this superiority translates into an advantage of AdamT during the early stage of the test loss.


\begin{figure*}
    \centering
        \includegraphics[height=2.4in]{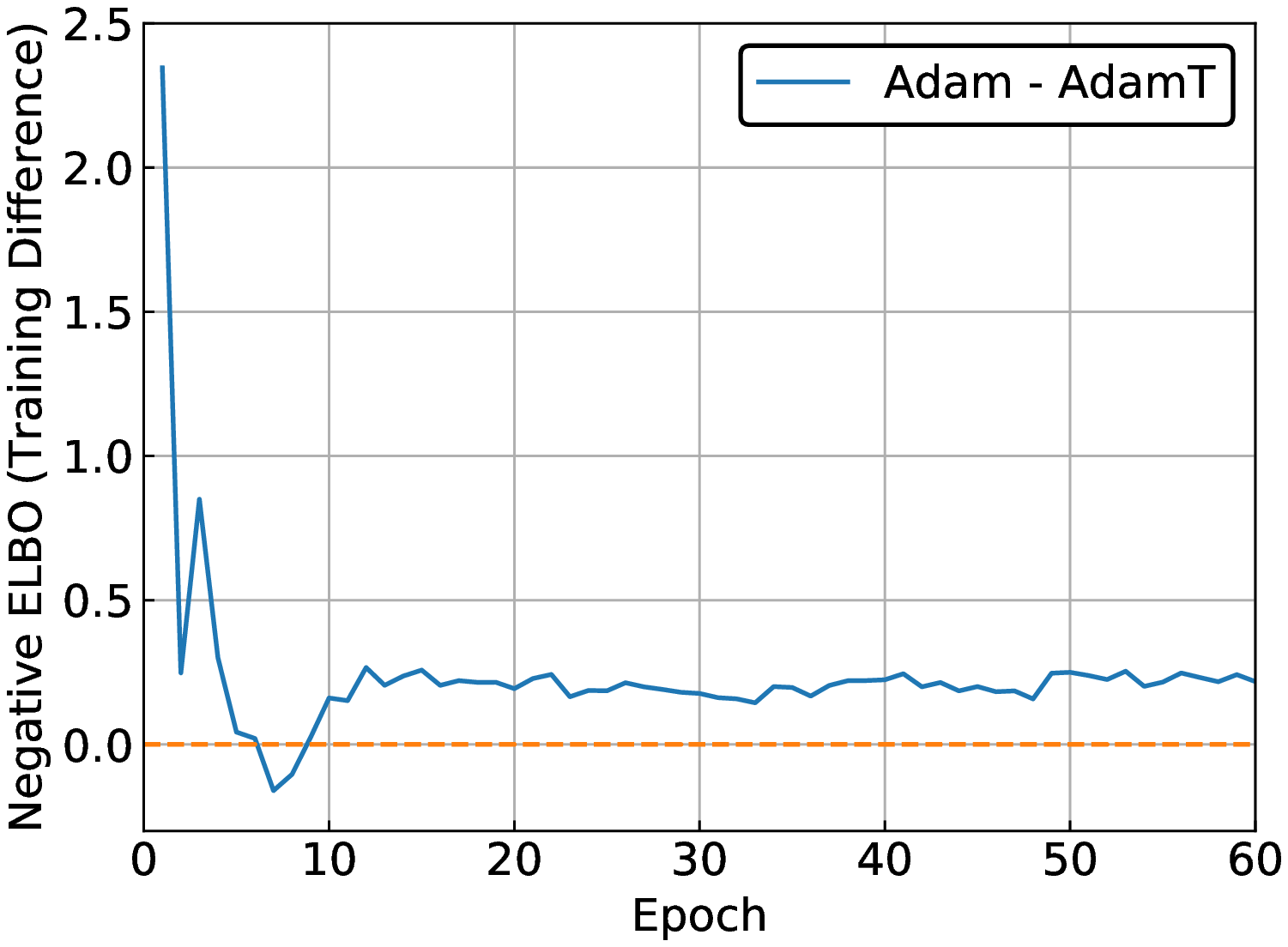}
        \includegraphics[height=2.4in]{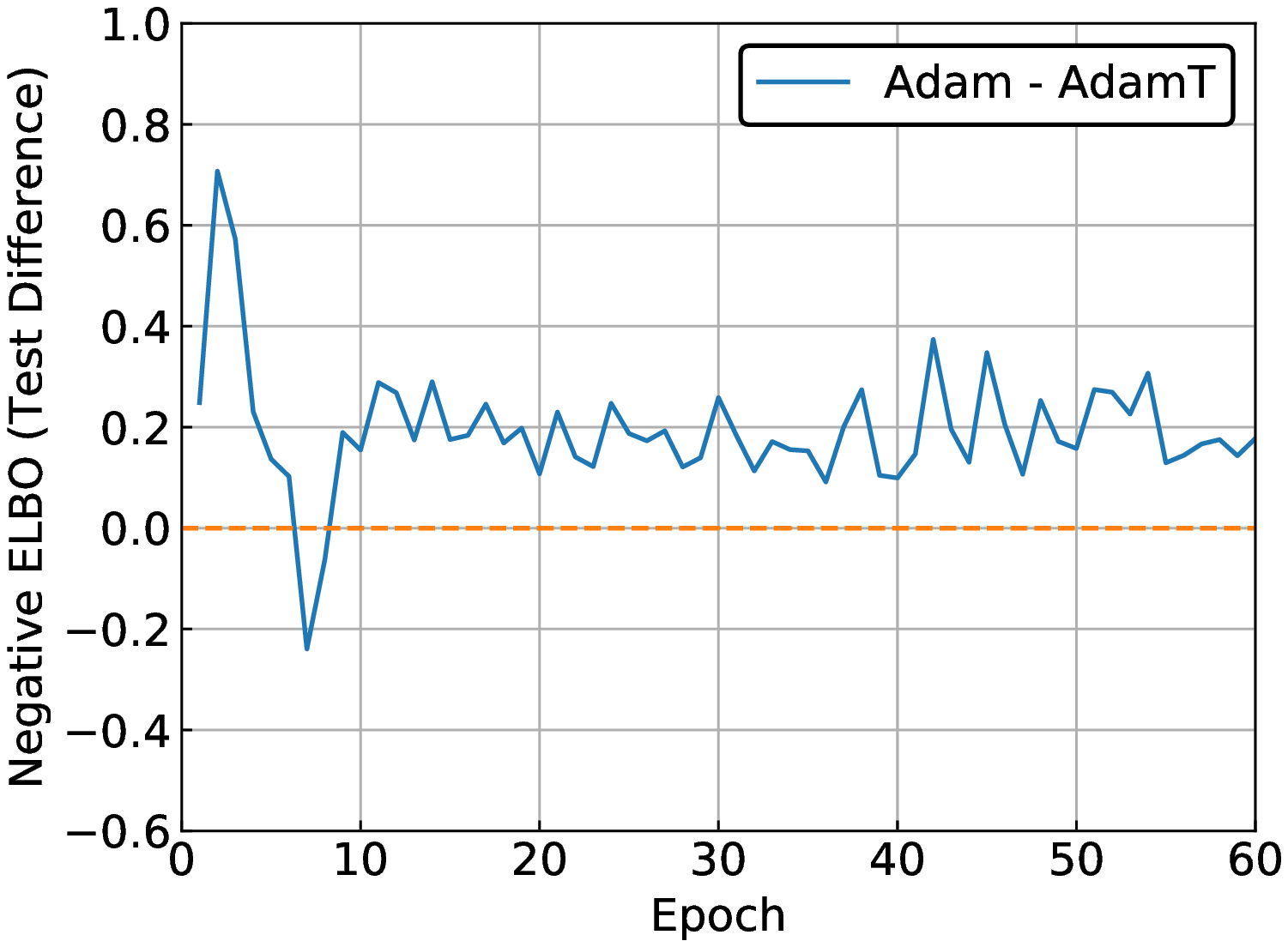}
    \caption{Training difference of negative ELBO (left) and test difference of negative ELBO (right) of the variational autoencoder on MNIST for density estimation.} 
    \label{exp_vae} 
\end{figure*}


\begin{center}
\begin{table*}[ht]
{\caption{The final training loss and test loss of each experiment. The reported numbers are the averages over $10$ repeated experiments with corresponding standard deviations.}\label{table1}}
\begin{center}
\begin{tabular}{c c c c c}
\hline
\rule{0pt}{8pt}
  & \multicolumn{2}{c}{Training Loss} & \multicolumn{2}{c}{Test Loss} \\
  \cline{2-5}
  & Adam & Adam+Trend & Adam & Adam+Trend 
\rule{0pt}{8pt}
\\
\hline
\\[-6pt]
LR & $0.3645\pm0.0005$ & $\mathbf{0.3634}\pm\mathbf{0.0003}$ & $0.4403\pm0.0028$ & $\mathbf{0.4390}\pm\mathbf{0.0025}$\\

FNN* & $0.1468\pm0.0036$ & $\mathbf{0.1386}\pm\mathbf{0.0018}$ & $\mathbf{0.7492}\pm\mathbf{0.0288}$ & $0.7639\pm0.0140$ \\

FNN & $0.5390\pm0.0046$ & $\mathbf{0.5173}\pm\mathbf{0.0045}$ & $0.5478\pm0.0066$ & $\mathbf{0.5376}\pm\mathbf{0.0073}$ \\

CNN & $0.2661\pm0.0039$ & $\mathbf{0.2488}\pm\mathbf{0.0034}$ & $\mathbf{0.8201}\pm\mathbf{0.0111}$ & $0.8228\pm0.0182$ \\

VAE & $244.1535\pm0.1202$ & $\mathbf{244.0036}\pm\mathbf{0.1398}$ & $245.9947\pm0.1996$ & $\mathbf{245.8023}\pm\mathbf{0.1600}$ \\
\hline\hline
\\[-6pt]
& AMSGrad & AMSGrad+Trend & AMSGrad & AMSGrad+Trend \\
\hline
\\[-6pt]
LR & $0.3686\pm0.0005$ & $\mathbf{0.3676}\pm\mathbf{0.0003}$ & $0.4390\pm0.0028$ & $\mathbf{0.4374}\pm\mathbf{0.0020}$\\

FNN* & $0.1622\pm0.0029$ & $\mathbf{0.1532}\pm\mathbf{0.0022}$ & $\mathbf{0.6982}\pm\mathbf{0.0186}$ & $0.7060\pm0.0142$ \\

FNN & $0.5419\pm0.0047$ & $\mathbf{0.5261}\pm\mathbf{0.0052}$ & $0.5475\pm0.0040$ & $\mathbf{0.5422}\pm\mathbf{0.0037}$ \\

CNN & $0.3055\pm0.0108$ & $\mathbf{0.2968}\pm\mathbf{0.0090}$ & $0.7974\pm0.0149$ & $\mathbf{0.7804}\pm\mathbf{0.0118}$ \\

VAE & $246.4323\pm0.1369$ & $\mathbf{246.3635}\pm\mathbf{0.1982}$ & $248.1823\pm0.1715$ & $\mathbf{248.1336}\pm\mathbf{0.2331}$ \\
\hline
\\[-6pt]
\end{tabular}
\end{center}
\end{table*}

\begin{table*}[ht]
\caption{The final training and test classification accuracy of each experiment. The reported numbers are the averages over $10$ repeated experiments with corresponding standard deviations.}\label{table2}
\begin{center}
\begin{tabular}{c c c c c}
\hline
\rule{0pt}{8pt}
  & \multicolumn{2}{c}{Training Accuracy} & \multicolumn{2}{c}{Test Accuracy} \\
  \cline{2-5}
  & Adam & Adam+Trend & Adam & Adam+Trend 
\rule{0pt}{8pt}
\\
\hline
\\[-6pt]
LR & $0.8732\pm0.0016$ & $\mathbf{0.8740}\pm\mathbf{0.0010}$ & $0.8063\pm0.0438$ & $\mathbf{0.8188}\pm\mathbf{0.0337}$ \\

FNN* & $0.9599\pm0.0058$ & $\mathbf{0.9625}\pm\mathbf{0.0037}$ & $\mathbf{0.8792}\pm\mathbf{0.0224}$ & $0.8771\pm0.0237$ \\

FNN & $0.8881\pm0.0027$ & $\mathbf{0.8933}\pm\mathbf{0.0033}$ & $0.8542\pm0.0349$ & $\mathbf{0.8646}\pm\mathbf{0.0104}$ \\

CNN & $0.9674\pm0.0001$ & $\mathbf{0.9703}\pm\mathbf{0.0011}$ & $ 0.6563\pm0.0313$ & $\mathbf{0.6875}\pm\mathbf{0.0625}$\\

VAE & - & - & - & - \\
\hline\hline
\\[-6pt]
& AMSGrad & AMSGrad+Trend & AMSGrad & AMSGrad+Trend \\
\hline
\\[-6pt]
LR & $0.8722\pm0.0015$ & $\mathbf{0.8732}\pm\mathbf{0.0010}$ & $0.8188\pm0.0337$ & $\mathbf{0.8250}\pm\mathbf{0.0250}$ \\

FNN* & $0.9551\pm0.0049$ & $\mathbf{0.9600}\pm\mathbf{0.0032}$ & $0.8958\pm0.0264$ & $\mathbf{0.9000}\pm\mathbf{0.0292}$ \\

FNN & $0.8834\pm0.0018$ & $\mathbf{0.8864}\pm\mathbf{0.0021}$ & $0.8458\pm0.0250$ & $\mathbf{0.8563}\pm\mathbf{0.0197}$ \\

CNN & $0.9529\pm0.0045$ & $\mathbf{0.9563}\pm\mathbf{0.0036}$ & $0.6688\pm0.0563$ & $\mathbf{0.6875}\pm\mathbf{0.0484}$\\

VAE & - & - & - & - \\
\hline
\\[-6pt]
\end{tabular}
\end{center}
\end{table*}

\end{center}


\subsection{Deep Generative Models For MNIST}
Variational Autoencoder (VAE) \cite{kingma2013auto,rezende2014stochastic} is one of the most popular deep generative models for density estimation and image generation. In this experiment, we train a VAE model on the standard MNIST dataset which contains $60,000$ training samples and $10,000$ test samples. Each sample is one $28\times 28$ black-and-white image of the handwritten digit. The VAE model used in this experiment exactly matches the architecture presented in \cite{kingma2013auto}: Gaussian encoder and Bernoulli decoder, both of which are implemented by feedforward neural networks with single hidden layer and there are $500$ hidden units in each hidden layer. We employ the hyperbolic tangent activation function for the model and set the dimensionality of the latent space as $20$. We use the constant learning rate and set the minibatch size to $128$.

We examine the Evidence Lower Bound (ELBO) of the training and test phases for the two optimizers. See Fig.~\ref{exp_vae} for the convergence results. Due to the scale issue, we plot the difference between the ELBOs produced by the two optimizers. Similar to the first experiment, we plot the difference value $(ELBO_{\textnormal{Adam}}-ELBO_{\textnormal{AdamT}})$ against the epoch for training and testing. We observe that our AdamT has a much faster convergence at the early stage of training than Adam and constantly excels Adam during the rest of the training phase. The superior performance of AdamT in this experiment also translates into a clear advantage in the test phase.

\subsection{Quantitative Evaluations}
We compare our proposed algorithms AdamT and AMSGradT with the corresponding baseline approaches Adam and AMSGrad respectively based on the final training loss, test loss, classification performance (except for the VAE model) in each experiment. The results recorded in Table~\ref{table1} and Table~\ref{table2} are the average values along with the standard deviations calculated over $10$ repeated experiments with random initializations. $\textnormal{FNN}^{*}$ denotes the feedforward neural networks without dropout layers while FNN represents the same model equipped with dropout layers. The results show that our proposed two trend-corrected variants have a superior performance over their corresponding baseline methods in most of the conducted experiments under different evaluation metrics, except for the models ($\textnormal{FNN}^{*}$ and CNN) which are prone to over-fitting. From the documented experiment results, we conclude that the additional trend estimates can provide meaningful information of the landscape of the loss surface and thus yield a better performance for the trend-corrected schemes.

{Table2}

\section{Related Works}\label{relatedWorK}
We consider the class of adaptive moment estimation methods with exponential moving average scheme as Adam-type learning algorithms. The fundamental idea was proposed in \cite{kingma2014adam} and quickly extended to many variants. Some examples include AdaMax \cite{kingma2014adam}, Nadam \cite{dozat2016incorporating} and AdamW \cite{loshchilov2018decoupled}. 

Despite the efficiency in practice, the problematic short-term memory of the gradients prevent the conventional Adam-type methods from a promising global convergences \cite{reddi2018convergence}. For the convex settings, they proposed AMSGrad that promises a global optimization with a comparable performance. Except for some other recent studies for convex optimization \cite{xu2017stochastic,chen2018sadagrad,levy2018online}, several works developed optimization methods for non-convex problems. Padam \cite{chen2018closing,zhou2018convergence} introduces a partial adaptive parameter to interpolate between SGD with momentum and Adam, so that adjacent learning rates could decrease smoothly. AdaUSM \cite{zou2018convergence} appends the idea of unified momentum for non-decreasing sequence of weights. AdaFom \cite{chen2019convergence} obtains first-order stationary by taking simple average on the second moment estimation. More conditions for pursuing global convergence were summarized in \cite{zou2019sufficient}, basing on the currently successful variants.


\section{Discussion and Conclusion}
In this work, we propose a new scheme to calculate the adaptive step size with trend-corrected exponential smoothing. The effectiveness of incorporating the trend information is investigated on the plain Adam method, which is the fundamental form of all Adam-type methods proposed in the literature as well as one of the most popular and widely used optimizers in practice with many irreplaceable advantages. On top of that, we also testify the contribution of the modified component on AMSGrad. Empirical results in Section \ref{experiment} demonstrate a performance gain on both optimizers in terms of convergence speed and robustness.

We leave some potentials for future developments. First, although we focused primarily on Adam, we believe that similar ideas could also be extended to other popular adaptive gradient methods for theoretical and experimental analysis
. Also, 
despite the computational feasibility, the initial implementation on Adam inherits its non-converge flaw. While this work focus on investigating the effectiveness of adding additional trend scheme instead of fixing the inherent issues, we try the modification with AMSGrad, as one of the solutions, to demonstrate the potential to extend our framework to optimizers that promise convergence on convex settings. 
For non-convex scenarios, some potential works in the recent literature are discussed in Section \ref{relatedWorK}.

\bibliographystyle{ieeetr}
\bibliography{ecai.bib}

\begin{thebibliography}{10}

\bibitem{duchi2011adaptive}
J.~Duchi, E.~Hazan, and Y.~Singer, ``Adaptive subgradient methods for online
  learning and stochastic optimization,'' {\em Journal of Machine Learning
  Research}, vol.~12, no.~Jul, pp.~2121--2159, 2011.

\bibitem{ruder2016overview}
S.~Ruder, ``An overview of gradient descent optimization algorithms,'' {\em
  preprint arXiv:1609.04747}, 2016.

\bibitem{tieleman2012lecture}
T.~Tieleman and G.~Hinton, ``Lecture 6.5-rmsprop: Divide the gradient by a
  running average of its recent magnitude,'' {\em COURSERA: Neural Networks for
  Machine Learning}, vol.~4, no.~2, pp.~26--31, 2012.

\bibitem{zeiler2012adadelta}
M.~D. Zeiler, ``Adadelta: an adaptive learning rate method,'' {\em preprint
  arXiv:1212.5701}, 2012.

\bibitem{kingma2014adam}
D.~P. Kingma and J.~Ba, ``{ADAM}: A method for stochastic optimization,'' in
  {\em Proceedings of International Conference on Learning Representation
  (ICLR)}, 2015.

\bibitem{qian1999momentum}
N.~Qian, ``On the momentum term in gradient descent learning algorithms,'' {\em
  Neural Networks}, vol.~12, no.~1, pp.~145--151, 1999.

\bibitem{reddi2018convergence}
S.~J. Reddi, S.~Kale, and S.~Kumar, ``On the convergence of {ADAM} and
  beyond,'' in {\em Proceedings of International Conference on Learning
  Representation (ICLR)}, 2018.

\bibitem{dozat2016incorporating}
T.~Dozat, ``Incorporating {Nesterov} momentum into {ADAM},'' in {\em
  Proceedings of 4th International Conference on Learning Representations,
  Workshop Track}, 2016.

\bibitem{loshchilov2018decoupled}
I.~Loshchilov and F.~Hutter, ``Decoupled weight decay regularization,'' in {\em
  International Conference on Learning Representations}, 2019.

\bibitem{liu2020on}
L.~Liu, H.~Jiang, P.~He, W.~Chen, X.~Liu, J.~Gao, and J.~Han, ``On the variance
  of the adaptive learning rate and beyond,'' in {\em International Conference
  on Learning Representations}, 2020.

\bibitem{holt2004forecasting}
C.~C. Holt, ``Forecasting seasonals and trends by exponentially weighted moving
  averages,'' {\em International Journal of Forecasting}, vol.~20, no.~1,
  pp.~5--10, 2004.

\bibitem{gardner1985forecasting}
E.~S. Gardner~Jr and E.~McKenzie, ``Forecasting trends in time series,'' {\em
  Management Science}, vol.~31, no.~10, pp.~1237--1246, 1985.

\bibitem{hyndman2018forecasting}
R.~J. Hyndman and G.~Athanasopoulos, {\em Forecasting: Principles and
  Practice}.
\newblock OTexts, 2018.

\bibitem{xiao2017fashion}
H.~Xiao, K.~Rasul, and R.~Vollgraf, ``Fashion-{MNIST}: a novel image dataset
  for benchmarking machine learning algorithms,'' {\em preprint
  arXiv:1708.07747}, 2017.

\bibitem{netzer2011reading}
Y.~Netzer, T.~Wang, A.~Coates, A.~Bissacco, B.~Wu, and A.~Y. Ng, ``Reading
  digits in natural images with unsupervised feature learning,'' {\em NIPS
  Workshop on Deep Learning and Unsupervised Feature Learning}, 2011.

\bibitem{srivastava2014dropout}
N.~Srivastava, G.~Hinton, A.~Krizhevsky, I.~Sutskever, and R.~Salakhutdinov,
  ``Dropout: a simple way to prevent neural networks from overfitting,'' {\em
  The Journal of Machine Learning Research}, vol.~15, no.~1, pp.~1929--1958,
  2014.

\bibitem{kingma2013auto}
D.~P. Kingma and M.~Welling, ``Auto-encoding variational bayes,'' in {\em
  Proceedings of International Conference on Learning Representations}, 2014.

\bibitem{rezende2014stochastic}
D.~J. Rezende, S.~Mohamed, and D.~Wierstra, ``Stochastic backpropagation and
  approximate inference in deep generative models,'' in {\em Proceedings of the
  31st International Conference on Machine Learning}, 2014.

\bibitem{xu2017stochastic}
Y.~Xu, Q.~Lin, and T.~Yang, ``Stochastic convex optimization: Faster local
  growth implies faster global convergence,'' in {\em Proceedings of the 34th
  International Conference on Machine Learning-Volume 70}, pp.~3821--3830,
  JMLR. org, 2017.

\bibitem{chen2018sadagrad}
Z.~Chen, Y.~Xu, E.~Chen, and T.~Yang, ``Sadagrad: Strongly adaptive stochastic
  gradient methods,'' in {\em International Conference on Machine Learning},
  pp.~912--920, 2018.

\bibitem{levy2018online}
Y.~K. Levy, A.~Yurtsever, and V.~Cevher, ``Online adaptive methods,
  universality and acceleration,'' in {\em Advances in Neural Information
  Processing Systems}, pp.~6500--6509, 2018.

\bibitem{chen2018closing}
J.~Chen and Q.~Gu, ``Closing the generalization gap of adaptive gradient
  methods in training deep neural networks,'' {\em preprint arXiv:1806.06763},
  2018.

\bibitem{zhou2018convergence}
D.~Zhou, Y.~Tang, Z.~Yang, Y.~Cao, and Q.~Gu, ``On the convergence of adaptive
  gradient methods for nonconvex optimization,'' {\em preprint
  arXiv:1808.05671}, 2018.

\bibitem{zou2018convergence}
F.~Zou and L.~Shen, ``On the convergence of weighted {AdaGrad} with momentum
  for training deep neural networks,'' {\em preprint arXiv:1808.03408}, 2018.

\bibitem{chen2019convergence}
X.~Chen, S.~Liu, R.~Sun, and M.~Hong, ``On the convergence of a class of
  {ADAM}-type algorithms for non-convex optimization,'' in {\em Proceedings of
  International Conference on Learning Representation (ICLR)}, 2019.

\bibitem{zou2019sufficient}
F.~Zou, L.~Shen, Z.~Jie, W.~Zhang, and W.~Liu, ``A sufficient condition for
  convergences of {ADAM} and {RMSProp},'' in {\em Proceedings of the IEEE
  Conference on Computer Vision and Pattern Recognition}, pp.~11127--11135,
  2019.

\end{thebibliography}

\end{document}